\documentclass{article}

\usepackage{arxiv}

\usepackage[utf8]{inputenc} % allow utf-8 input
\usepackage[T1]{fontenc}    % use 8-bit T1 fonts
\usepackage{hyperref}       % hyperlinks
\usepackage{url}            % simple URL typesetting
\usepackage{booktabs}       % professional-quality tables
\usepackage{amsfonts}       % blackboard math symbols
\usepackage{nicefrac}       % compact symbols for 1/2, etc.
\usepackage{microtype}      % microtypography
\usepackage{lipsum}		% Can be removed after putting your text content
\usepackage{graphicx}
\usepackage{natbib}
\usepackage{doi}
\usepackage{amsmath}
\usepackage{comment}

\title{Optimizing 4D Wires for Sparse 3D Abstraction}

\author{ \href{https://orcid.org/0000-0000-0000-0000}{\includegraphics[scale=0.06]{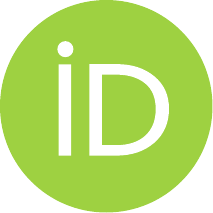}\hspace{1mm}Dong-Yi Wu}\thanks{Use footnote for providing further
		information about author (webpage, alternative
		address)---\emph{not} for acknowledging funding agencies.} \\
	National Cheng Kung University\\
	Tainan, Taiwan \\
	\texttt{cutechubbit@gmail.com} \\
	\And
	\href{https://orcid.org/0000-0000-0000-0000}{\includegraphics[scale=0.06]{orcid.pdf}\hspace{1mm}Tong-Yee Lee} \\
	National Cheng Kung University\\
	Tainan, Taiwan \\
	\texttt{tonylee@mail.ncku.edu.tw} \\
}

\renewcommand{\shorttitle}{\textit{arXiv} Template}

\hypersetup{
pdftitle={A template for the arxiv style},
pdfsubject={q-bio.NC, q-bio.QM},
pdfauthor={David S.~Hippocampus, Elias D.~Striatum},
pdfkeywords={First keyword, Second keyword, More},
}

\begin{document}
\maketitle

\begin{abstract}
We present a unified framework for 3D geometric abstraction using a single continuous 4D wire, parameterized as a B-spline with spatial coordinates and variable width $(x,y,z,w)$. Existing approaches typically represent shapes as collections of many independent curve segments, which often leads to fragmented structures and limited physical realizability. In contrast, we show that a single continuous spline is sufficiently expressive to capture complex volumetric forms while enforcing global topological coherence. By imposing continuity, our method transforms 3D sketching from a local density-accumulation process into a global routing problem, providing a strong inductive bias toward cleaner aesthetics and improved structural coherence. To enable gradient-based optimization, we introduce a differentiable rendering pipeline that efficiently rasterizes variable-width curves with bounded projection error. This formulation supports robust optimization using modern guidance signals such as Score Distillation Sampling (SDS) or CLIP. We demonstrate applications including image-to-3D abstraction, multi-view wire art generation, and differentiable stylized surface filling. Experiments show that our unified representation produces structures with higher semantic fidelity and improved structural coherence compared to approaches based on collections of discrete curves.
\end{abstract}

% keywords can be removed

\section{Introduction}The abstraction of complex geometry into sparse, expressive lines has long been a central pursuit in both computer graphics and traditional art. With the recent surge of generative 3D techniques, significant progress has been made in text-to-3D and image-to-3D sketching. Methods based on Score Distillation Sampling (SDS) and differentiable rendering can now distill 3D shapes into sparse curve-based representations directly from text prompts or images. However, state-of-the-art approaches~\cite{dreamwire,3doodle,diff3ds} suffer from a fundamental limitation: they typically represent geometry as a fragmented collection of independent Bézier segments—effectively a “bag of curves.” While this representation offers local flexibility, it has several drawbacks (as visualized in Fig. \ref{fig:overview}): (1) \textbf{Lack of Structural Coherence.} The resulting geometry often consists of many floating curve segments rather than a unified structure. As strokes move independently, the distribution of geometry can become uneven, with some regions overly dense while others remain sparse or disconnected. Such inconsistencies reduce aesthetic clarity and limit physical realizability in applications such as wire bending or 3D fabrication. (2) \textbf{Limited Expressiveness of Local Primitives.} Individual Bézier segments provide only limited representational capacity. Capturing complex silhouettes or volumetric structure therefore requires a large number of short segments, which collectively approximate the target geometry. This reliance on many small primitives often produces visually noisy results compared to the smooth, continuous paths commonly found in human-drawn sketches. (3) \textbf{Fixed and Redundant Topology.} Redundant or erroneous strokes may persist once introduced, as the optimizer lacks effective mechanisms for merging or restructuring curve segments. This can lead to suboptimal solutions during optimization.

%%%%%%%%%%%%%%%%%%%%%%%%%%%

In this paper, we question the necessity of fragmented curve representations and investigate whether complex 3D structures can instead be abstracted using a single continuous primitive. Specifically, we extend the concept of continuous B-splines, which have proven effective for 2D vector graphics~\cite{2dbspline}, to the 3D domain and investigate the following question: \textit{Is a single continuous 1D manifold expressive enough to abstract complex 3D geometry?} While representing a 2D image with a single line is relatively intuitive, extending this idea to 3D is considerably more challenging, as the curve must encode the spatial organization of a volumetric structure rather than simply tracing planar contours. To address this challenge, we introduce a novel representation based on a single \textbf{4D Wire}. Rather than optimizing a large collection of short and independent curve segments, our framework initializes a single global B-spline that spans the entire structure and optimizes it through a Differentiable Vector Graphics (DiffVG \cite{diffvg}) pipeline. The curve is parameterized in four dimensions by jointly optimizing spatial coordinates $(x,y,z)$ and a continuous width function $w(t)$ along the curve.

We argue that optimizing stroke width is not merely an aesthetic enhancement but a geometric necessity for 3D abstraction. In 2D vector graphics, thick strokes can be approximated by stacking multiple thin curves. However, extending this strategy to 3D leads to dense bundles of curves that increase optimization complexity and introduce redundant geometry without improving structural expressiveness. By introducing a continuous width parameter $w(t)$, a single primitive can represent both volumetric mass through thicker regions and fine structural detail through thinner regions within a unified representation.

To render this 4D wire under perspective projection, we introduce an efficient differentiable rendering pipeline built on DiffVG~\cite{diffvg}. By deriving a bound on the projection error, we avoid the need for specialized rational curve rasterization while maintaining stable gradients for optimization. 

We demonstrate that this unified representation enables several applications that benefit from improved structural coherence and connectivity compared to fragmented curve collections:

\begin{itemize}

\item \textbf{Image-to-3D Stroke.} By leveraging Zero-1-to-3~\cite{zero123} as a multi-view generative prior, we distill the photometric information of a 2D image into a sparse 3D abstraction. The method lifts the semantic and geometric essence of a source image into a single, continuous 3D curve structure.

\item \textbf{Multi-view Wire Art.} We optimize a unified 4D wire under the guidance of 2D diffusion priors to synthesize artistic wire sculptures that exhibit view-dependent semantics.

\item \textbf{Semantic Surface Filling.} Starting from an input mesh, we optimize surface-filling curves to adopt semantic motifs—such as floral patterns or Chinese calligraphy. This process effectively abstracts the 3D volume into a continuous, stylized wireframe that reflects both the underlying geometry and high-level artistic concepts.
\end{itemize}

In summary, our contributions are:

\begin{itemize}
\item We show that a single continuous 4D wire is sufficiently expressive to abstract complex 3D geometry while maintaining topological continuity.

\item We introduce a differentiable rendering framework that extends DiffVG to efficiently rasterize variable-width 3D B-splines with bounded projection error.

\item We showcase the versatility of our representation across various 3D wire generation tasks, consistently yielding results with superior visual clarity and global connectivity that overcome the inherent structural fragmentation of discrete-curve parameterizations.
\end{itemize}

\section{Related Work}

\subsection{Differentiable Vector Graphics}

Differentiable vector graphics has recently emerged as a powerful framework for optimizing parametric vector primitives under image-based supervision. A key milestone in this area is DiffVG~\cite{diffvg}, which introduced a differentiable rasterization pipeline capable of propagating gradients from pixel space back to vector parameters such as control points, stroke widths, and opacities. By enabling gradient-based optimization over vector primitives, DiffVG established a new paradigm for learning and generating vector graphics directly from image-space objectives. 

One important application of differentiable vector graphics is raster-to-vector reconstruction. In this setting, vector primitives such as Bézier curves or parametric shapes are optimized to approximate raster images, producing scalable and editable representations. Recent works demonstrate that complex visual structures can be reconstructed through gradient-based optimization over a relatively small set of primitives~\cite{live, diffvg}. These approaches are particularly effective for curve-based abstraction and stylized vector graphics.

Another prominent direction focuses on sketch abstraction and generative drawing. Early approaches leveraged the semantic capabilities of CLIP~\cite{radford2021learning} to guide stroke placement. CLIPDraw~\cite{frans2022clipdraw} demonstrated text-to-SVG generation by optimizing strokes to match textual prompts in CLIP space. Building on this idea, CLIPasso~\cite{vinker2022clipasso} explored semantic sketch abstraction, producing simplified drawings that preserve salient image features while suppressing high-frequency details.

More recently, the introduction of Score Distillation Sampling (SDS)~\cite{dreamfusion} enabled the distillation of generative priors from diffusion models into vector representations. VectorFusion~\cite{jain2023vectorfusion} pioneered SDS-based text-to-SVG generation by optimizing vector primitives under supervision from a frozen Stable Diffusion model. Subsequent works expanded this paradigm to additional design tasks, including semantic typography~\cite{font} and improved sketch generation through attention-guided optimization~\cite{xu2023diffsketcher}. 

Despite these advances, existing approaches are primarily confined to 2D planar space. In contrast, our work extends differentiable vector optimization into the 3D domain. This introduces a fundamentally harder problem, as it requires optimizing a primitive across a vastly larger and more complex 3D solution space.

\subsection{Deep 3D Sketching and Abstraction}

While differentiable vector graphics have matured for planar imagery, extending these capabilities to the 3D domain introduces additional challenges such as occlusion handling and multi-view consistency. Recent works have begun to explore the optimization of 3D curve primitives for generative sketching. 3Doodle~\cite{3doodle} represents objects using collections of 3D cubic Bézier curves, optimized through differentiable rendering to approximate target geometry from multiple viewpoints. Diff3DS~\cite{diff3ds} further advances this direction by introducing a differentiable rendering framework for rational Bézier curves, enabling high-fidelity text-to-3D and image-to-3D sketch generation. Beyond general object abstraction, specialized applications such as anamorphic wire art have also been explored. DreamWire~\cite{dreamwire} leverages diffusion priors to optimize curve arrangements that produce specific projections from multiple viewpoints, effectively solving a multi-view shadow art problem.

\paragraph{Limit of Fragmentation.} Despite their visual success, these methods share a fundamental limitation: they represent geometry as a "bag of curves." By spawning and optimizing a disconnected collection of independent Bézier segments, they prioritize local visual density over global topological coherence. This fragmentation results in "floating" geometry that lacks physical structural integrity. In contrast, our unitary 4D wire formulation enforces strict mathematical continuity, transforming the task from a stochastic curve collection problem into a true topology optimization problem.

\subsection{Wire Art and Continuous Curve Modeling}
Beyond visual approximation using 3D curves, a distinct class of methods focuses on physical wire modeling, where topological \textit{connectivity} is a fundamental constraint.

\paragraph{Wire Abstraction and Fabrication.}
Several works generate wire representations by simplifying 3D geometry or optimizing for physical constraints. \citet{lab_wire_art} proposed an iterative abstraction method that extracts representative curves through mesh edge merging. To ensure physical realizability, \citet{planar_rod} investigated the construction of stable sculptures by computing structural equilibrium and layering order, while \citet{eulerian_wire} optimized Eulerian traversals to obtain minimal continuous structures. Recent toolkits like WireBend-kit~\cite{wire_bending} have further bridged this gap by providing hardware-software pipelines for converting 3D wireframes into robotic bending instructions. However, these methods primarily prioritize mechanical validity and reconstruction of existing forms, whereas our framework focuses on the generative synthesis of expressive, semantic-driven abstractions.

\paragraph{Reconstruction and Style Transfer.}
Wire geometry can also be recovered from image observations or 2D inputs. Methods such as \citet{image_wire_art} and Vid2Curve~\cite{vid2curve} reconstruct 3D curves from multi-view sequences using Structure-from-Motion (SfM), while WrapIt~\cite{wrapit} decomposes 2D line art into graph structures to infer fabricable wire compositions. While these approaches are effective for recovering existing thin structures or lifting planar drawings into 3D, they remain bound to the topology of the input observation. In contrast, our method leverages diffusion priors to synthesize wire art that does not yet exist, enabling complex applications like multi-view illusions and semantic surface ornamentation that go beyond simple geometric reconstruction.

\paragraph{Our Contribution.}
Our work bridges the domains of generative sketching and physical wire modeling. We introduce a \textbf{4D Wire} representation that enforces global topological continuity while enabling expressive geometric abstraction through variable-width curves. In contrast to existing generative sketching approaches that rely on collections of independent strokes, our method synthesizes a single continuous structure directly from semantic guidance, producing results that are both structurally coherent and visually expressive.

\section{The Unitary 4D Wire Primitive}
\begin{figure*}[t]
  \centering
  \includegraphics[width=\linewidth]{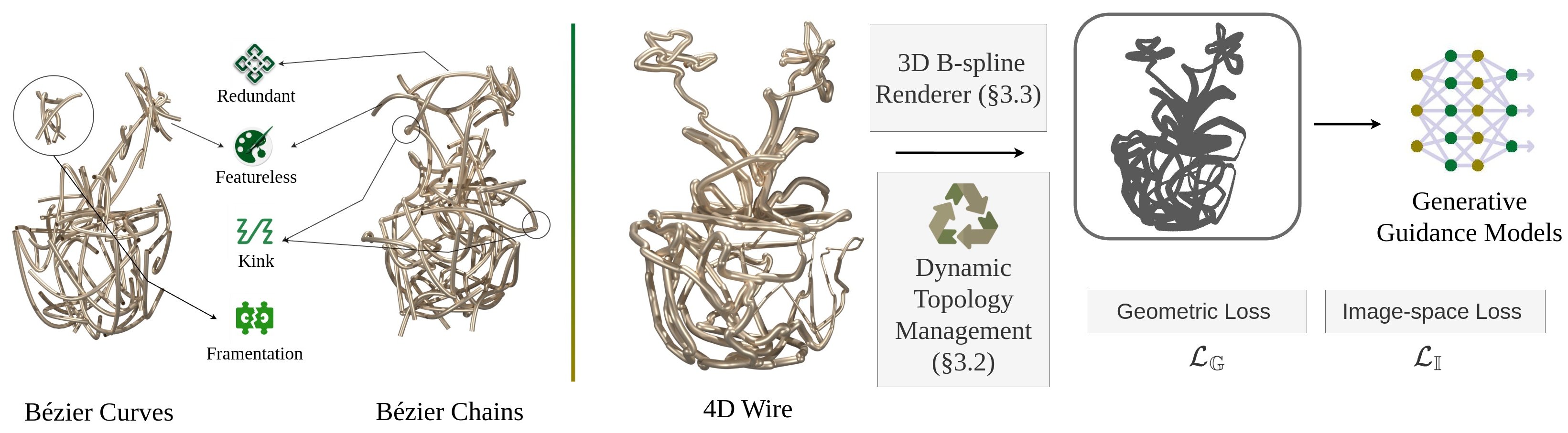}
  \caption{We compare our single 4D Wire representation against baseline approaches based on independent Bézier segments and Bézier chains. All models are optimized under identical generative guidance from the Zero123 \cite{zero123} diffusion prior and use an equivalent number of degrees of freedom. Our representation mitigates the structural fragmentation inherent in independent segments and avoids the angular discontinuities typical of standard $C^0$ Bézier chains. As a result, the 4D Wire achieves greater expressive power with fewer redundant artifacts, producing a more coherent 3D abstraction.}
  \label{fig:overview}
\end{figure*}

We provide an overview of our method in Fig.~\ref{fig:overview}. We begin by contrasting our 4D Wire representation with commonly used curve primitives, such as Bézier curves and Bézier chains. Our 4D Wire is optimized using two complementary objectives: an image-space loss and a geometric regularization loss. To integrate with pretrained generative models, the 4D Wire is rendered into images through a differentiable renderer, enabling gradient-based supervision from image-space guidance. Finally, our topology management strategy reduces redundancy and mitigates the lack of details observed in existing methods, resulting in more expressive and consistent abstractions.

\subsection{Mathematical Formulation}
We define the 4D Wire as a single parameterized curve $\mathcal{C}(t) = [x(t), y(t), z(t), w(t)]^\top$. The state of the curve for $t \in [t_{min}, t_{max}]$ is determined by $n+1$ control points $\mathbf{P}_i \in \mathbb{R}^4$ and a clamped uniform knot vector 
$U=\{u_0,u_1,\dots,u_m\}$.:

\begin{equation}
\mathcal{C}(t) = \sum_{i=0}^{n} \mathbf{P}_i B_{i,d}(t),
\end{equation}

where $B_{i,d}(t)$ denotes the B-spline basis functions of degree $d$. To balance geometric smoothness with optimization efficiency, we specifically utilize \textbf{uniform cubic B-splines} ($d=3$), which guarantee $C^2$ continuity across the entire wire. We favor cubic B-splines over higher-degree variants (such as quintic splines) for two primary reasons. First, they provide greater local control and flexibility, avoiding the over-stiffening effects that higher-degree polynomials often exhibit during complex geometric optimization. Second, uniform cubic B-splines can be seamlessly and analytically converted into sequences of cubic Bézier segments via a constant basis transformation matrix. This mathematical property is critical, as it ensures native compatibility with modern differentiable rendering engines (e.g., DiffVG) that inherently rely on cubic Bézier primitives, allowing the curve to be evaluated and differentiated rapidly without approximation errors.

To mimic realistic movement and ensure organic curvature, we apply a third-order smoothness constraint, or jerk loss \cite{2dbspline}:
\begin{equation}
\mathcal{L}_{jerk} = \int_{t_{min}}^{t_{max}} \left\| \frac{d^3}{dt^3} \mathcal{C}(t) \right\|^2 dt.
\end{equation}
This loss is applied across all four dimensions, ensuring that both spatial transitions and width variations remain smooth.

\subsection{Dynamic Topology Management}

A major limitation of existing curve-based optimization methods is the lack of explicit topology management. Once a curve primitive is introduced, it typically persists throughout optimization—it can be adjusted in position or shape, but cannot be removed even if it becomes redundant or leads to suboptimal solutions. Prior work attempts to alleviate this issue using heuristics such as periodic parameter resets or length-based pruning~\cite{dreamwire, diff3ds}. However, these strategies are often limited in their effectiveness and do not fully address the underlying redundancy. In contrast, we propose a topology management approach based on the 4D Wire representation that dynamically reallocates geometric capacity during optimization. Our method enables the model to adaptively refine, suppress, or reinitialize portions of the curve, leading to more efficient and expressive representations.

\subsubsection{Width-Guided Reinitialization} The 4D wire representation offers a unique optimization advantage unavailable to current curves optimization methods: the learnable stroke width $w$ serves as a dual-purpose variable, acting simultaneously as an expressive rendering attribute and a differentiable saliency signal. Conventional length-based pruning \cite{diff3ds,dreamwire} and reinitialization are often ineffective because geometric redundancy does not strictly imply zero arc-length in gradient-based optimization. A redundant curve segment will rarely shrink to a single point because doing so requires the coordinated spatial displacement of multiple control points against local gradient signals. In contrast, width is a purely localized parameter. If a segment is visually redundant, the optimizer can independently drive its width to zero ($w \to 0$) without altering the curve's spatial trajectory. 

We formalize this as a direct reinitialization criterion. Let $w_i$ denote the optimized width component of the $i$-th control point $\mathbf{P}_i$. We define the set of pruned control points $\mathcal{P}_{prune}$ as those whose width has vanished:

\begin{equation}
\mathcal{P}_{prune} = \left\{ \mathbf{P}_i \ \middle|\ w_i = 0 \right\}
\end{equation}

For any control point $\mathbf{P}_i \in \mathcal{P}_{prune}$, we randomly reinitialize its spatial coordinates and reset its width. This structural reset serves a critical dual purpose. First, it acts as an active recycling mechanism, reclaiming the model's finite geometric capacity from visually redundant segments ($w = 0$). Second, by reinitializing these points and introducing sudden topological changes to the rendered image, the reset injects a discrete perturbation into the optimization process. This perturbation forces the model to break out of stagnant local minima, encouraging the emergence of new structural features while strictly preserving the continuous connectivity of the 4D wire.
\begin{figure*}[]
  \centering
  \includegraphics[width=0.9\linewidth]{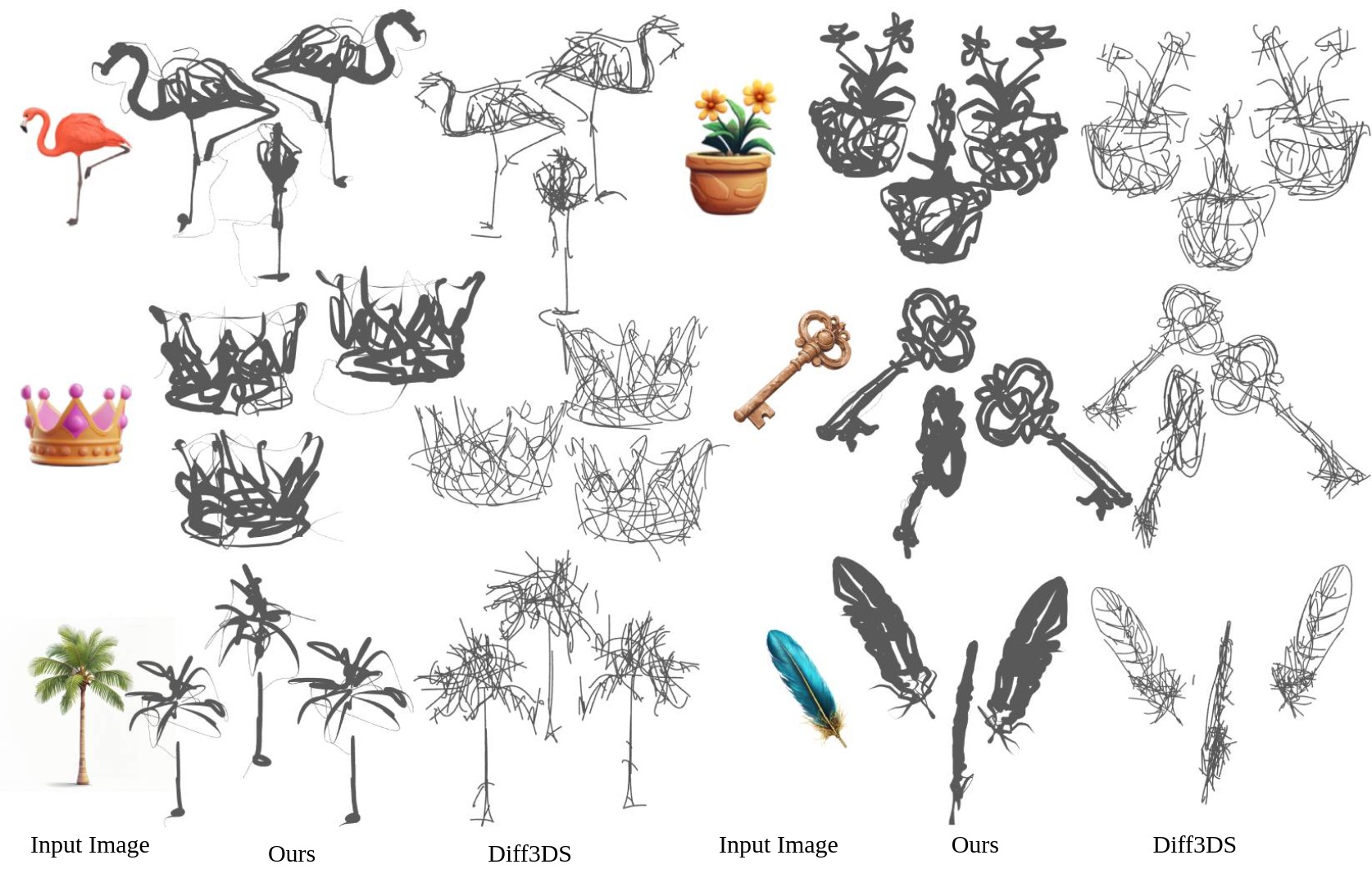}
  \caption{We compare the reconstructed 3D strokes generated from a single input image (left). Utilizing an equivalent parameter budget, our 4D Wire yields structurally unified results that better abstract the semantic features of the target object compared to the discrete baseline.}
  \label{fig:comparison_diff3ds}
\end{figure*}

\subsubsection{Wire Refinement via Knot Insertion}
To adaptively increase geometric capacity in high-gradient regions, we employ gradient-based knot insertion. We identify segments where large control point gradients indicate that the current representation is insufficient to capture emerging structural details. Unlike standard Bézier subdivision, which introduces additional segments that may be optimized independently and can weaken continuity constraints (in practice often leading to visible kinks at segment boundaries), knot insertion (e.g., Boehm's algorithm~\cite{knot_insertion}) preserves the global B-spline formulation. By inserting a new knot $\hat{u}$ into the knot vector $U$, we obtain a refined set of control points $\hat{\mathbf{P}}$ such that the resulting curve $\hat{\mathbf{C}}(t)$ remains geometrically identical to the original $\mathbf{C}(t)$:
\begin{equation}
\hat{\mathbf{C}}(t) \equiv \mathbf{C}(t), \quad \forall t.
\end{equation}

Importantly, the refined control points remain globally coupled through the B-spline basis, ensuring that the original continuity of the curve (e.g., $C^2$ for cubic splines) is preserved during subsequent optimization. This enables localized refinement while maintaining smoothness and avoiding artificial discontinuities.

\subsection{Differentiable Perspective Rendering}

%<<<<<<<<,,,
To integrate with differentiable vector graphics engines such as DiffVG~\cite{diffvg}, which natively support only polynomial 2D Bézier curves, we must bridge the gap between 3D B-spline geometry and 2D planar rendering. A fundamental challenge arises from perspective projection: due to the depth-dependent division term $1/z(t)$, the projection of a polynomial 3D B-spline is analytically a 2D rational curve (i.e., a NURBS representation). 

Recent approaches such as Diff3D~\cite{diff3ds} address this by introducing custom rational rasterizers. However, explicitly evaluating the rational denominator incurs significant computational overhead in both the forward pass and gradient backpropagation. 

Instead, we propose a dense basis conversion strategy that avoids explicit rational evaluation, enabling compatibility with standard, highly optimized polynomial rasterizers. Our key insight is that the projection error between the true rational curve and its polynomial approximation decreases rapidly as the segment length becomes small. By upsampling the sparse 3D B-spline into a dense sequence of 3D Bézier segments prior to projection, we obtain an accurate approximation of the true projection while eliminating the need for rational evaluation.

We formulate this conversion as a single global linear operation. Given $n+1$ sparse cubic B-spline control points $\mathbf{P}_{\text{sparse}} \in \mathbb{R}^{(n+1) \times 3}$, we compute a dense set of $m$ Bézier control points $\mathbf{P}_{\text{dense}} \in \mathbb{R}^{m \times 3}$ via:
\begin{equation}
\mathbf{P}_{\text{dense}} = \mathbf{M}_{\text{dense}} \, \mathbf{P}_{\text{sparse}},
\end{equation}
where the global transformation matrix $\mathbf{M}_{\text{dense}} \in \mathbb{R}^{m \times (n+1)}$ combines basis conversion and subdivision:

\begin{equation}
\mathbf{M}_{\text{dense}} =
\underbrace{
\begin{bmatrix}
\mathbf{S}_0 \\
\mathbf{S}_1 \\
\vdots \\
\mathbf{S}_{N-1}
\end{bmatrix}
}_{\text{Subdivision Stack} \in \mathbb{R}^{m \times k}}
\;
\underbrace{
\mathbf{M}_{B \to Z}
}_{\text{Basis Conversion} \in \mathbb{R}^{k \times (n+1)}}.
\end{equation}

%%%<<<<<<<<<<<<<<<<<<
Here, $\mathbf{M}_{B \to Z}$ converts the uniform cubic B-spline basis into the equivalent cubic Bézier basis~\cite{spline2bezier}, producing $k$ intermediate Bézier control points. The subdivision stack consists of matrices $\mathbf{S}_i$ that encode De Casteljau subdivision~\cite{subdivision}, generating a dense sequence of control points across $N$ segments.

By refining the curve into sufficiently small segments, the approximation error between the true projected curve $\mathbf{R}(s)$ and its polynomial approximation $\mathbf{L}(s)$ is bounded by:
\begin{equation}
\| \mathbf{R}(s) - \mathbf{L}(s) \| \le \mathcal{O}\left( \frac{h^2}{z_{\min}^2} \right),
\end{equation}
where $h$ denotes the segment length and $z_{\min}$ is the minimum depth along the curve. We provide the full derivation in the supplemental material. This bound implies that the approximation error scales with $\left(\frac{h}{z_{\min}}\right)^2$, yielding two practical insights: (1) quadratic convergence, where reducing the segment length $h$ rapidly decreases projection error, and (2) an adaptive subdivision strategy, where segments near the camera (small $z$) require finer subdivision to maintain a consistent error bound.

\begin{figure}[]
  \centering
  \includegraphics[width=0.8\linewidth]{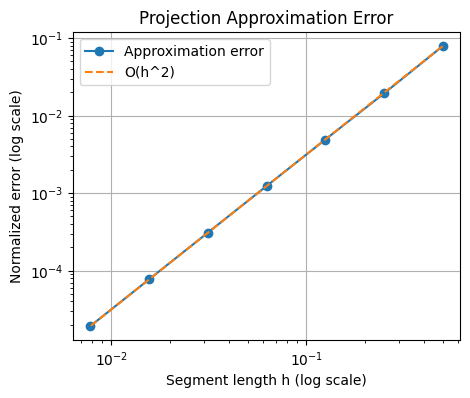}
  \caption{We measure the screen-space error between the projected reference curve, obtained via dense sampling with exact perspective division, and our polynomial approximation at varying subdivision levels. The error is normalized by the projected 2D bounding-box diagonal to ensure scale invariance. The log--log plot shows that the approximation error (blue) decreases approximately quadratically with segment length $h$, consistent with the theoretical bound in Eq.~(7) (orange dotted line). At the subdivision level used in our method, the error is negligible in practice.}
  \label{fig:error}
\end{figure}

To empirically validate this approximation, we compare our approximation against a high-resolution reference obtained by densely sampling the original 3D curve and projecting the samples using exact perspective division. To ensure comparability across camera settings and object scales, we normalize all curves to unit 3D bounding-box diagonal and report error in screen space normalized by the projected 2D bounding-box diagonal. We measure the approximation error as the mean Euclidean distance between corresponding points on the two curves in screen space, where both curves are uniformly resampled with respect to arc length. We evaluate this error under increasing subdivision levels, where shorter segments correspond to smaller approximation step size $h$. As shown in Fig.~\ref{fig:error}, the error decreases approximately quadratically with respect to $h$, consistent with the bound in Eq.~(7). In practice, the subdivision level used in our rendering pipeline achieves sub-pixel accuracy in screen space while maintaining efficient computation.

Finally, because the conversion is a static linear transformation, the gradient flow is deterministic and efficient. The gradients with respect to the original sparse parameters are computed via the chain rule without topological overhead:
\begin{equation}\frac{\partial \mathcal{L}}{\partial \mathbf{P}_{\text{sparse}}} = \mathbf{M}_{\text{dense}}^T \cdot \frac{\partial \mathcal{L}}{\partial \mathbf{P}_{\text{dense}}}.\end{equation} 

\paragraph{Implementation.}
The practical rendering pipeline consists of three stages:
(1) \textbf{Exact 3D Basis Conversion}: Using $\mathbf{M}_{\text{dense}}$, the sparse 3D B-spline is algebraically converted into a dense set of explicit 3D Bézier segments. This step is geometrically exact in $\mathbb{R}^3$ and preserves the original curve topology.
(2) \textbf{Projective Transformation}: The dense 3D Bézier control points are projected into screen space using the camera intrinsics. Because the segments are sufficiently short, we treat the projected points as standard 2D Bézier control points, implicitly approximating the rational weight as locally constant.
(3) \textbf{Polynomial Rasterization}: The resulting 2D polynomial Bézier paths are directly rasterized using DiffVG.

Overall, this linearized rendering pipeline achieves a $2$--$3\times$ speedup in wall-clock time compared to rational rasterization baselines, while maintaining sub-pixel accuracy.

\section{Applications}In this section, we evaluate the capacity of the a single 4D Wire representation by applying it to three distinct 3D curve generation tasks: image-to-3D stroke generation (Section 4.1), multi-view wire art synthesis (Section 4.2) and the optimization of differentiable stylized surface-filling curves (Section 4.3).

\subsection{Image-to-3D Stroke} 

The objective of image-to-3D stroke reconstruction is to distill the volumetric essence of an object into a sparse and abstract 1D topology. Current state-of-the-art methods, such as Diff3DS~\cite{diff3ds}, optimize an unstructured collection of Bézier curves guided by the Zero-1-to-3~\cite{zero123} generative prior through Score Distillation Sampling (SDS). In contrast, we employ our 4D Wire representation for the same task (Fig.~\ref{fig:comparison_diff3ds}).

%%<<<<<<<<<<<<<<<<<<<<<<<<
\paragraph{Initialization.}
Given a single reference image $I_{\text{ref}}$, we leverage the object's 2D silhouette to construct an initialization for the 3D curve optimization. Specifically, we define a bounded volume $\mathcal{V} \subset \mathbb{R}^3$ using the visual hull~\cite{visual_hull} derived from the silhouette mask $M_{\text{ref}}$. To obtain a continuous and non-intersecting topology, we sample a set of points $P_{\text{init}} \subset \mathcal{V}$ and solve a 3D Euclidean Traveling Salesperson Problem (TSP) over these points. The resulting path forms a single continuous trajectory that spans the visual hull and serves as the initialization of the B-spline control points. This strategy can be viewed as a 3D extension of TSP-based initialization used in prior 2D vector abstraction methods~\cite{2dbspline}.

\paragraph{Optimization.}
In our experiments, we use 100--220 control points for the 4D Wire and optimize for 900 iterations using the Adam optimizer. We observe that using an excessive number of control points can lead to cluttered structures. To encourage exploration during early optimization, we apply \emph{width-guided reinitialization} at iterations 150 and 300. For final detail enhancement, we perform a \emph{wire refinement} step at iteration 600. 

The optimization objective consists of an image-space semantic loss $\mathcal{L}_{\mathbb{I}}$ and a geometric regularization loss $\mathcal{L}_{\mathbb{G}}$:
\begin{equation}
\mathcal{L}_{\text{total}} = \lambda_{\mathbb{I}} \mathcal{L}_{\mathbb{I}} + \lambda_{\mathbb{G}} \mathcal{L}_{\mathbb{G}}.
\end{equation}

\paragraph{Image Space Loss ($\mathcal{L}_{\mathbb{I}}$).}
We enforce multi-view semantic consistency using Score Distillation Sampling (SDS) derived from Stable Zero123~\cite{stable_zero123}. Given a rendered image $I = \mathcal{R}(\tilde{\Theta})$ from a camera pose with extrinsics $(R, T)$, the SDS gradient updates the curve parameters $\tilde{\Theta}$ to align the rendering with the diffusion prior:
\begin{equation}
\nabla_{\tilde{\Theta}}\mathcal{L}_{\mathbb{SDS}}(\phi, I) =
\mathbb{E}_{t,\epsilon}
\left[
\omega(t)\left(\hat{\epsilon}_{\phi}(I_t; \tilde{I}, R, T, t) - \epsilon\right)
\frac{\partial I}{\partial \tilde{\Theta}}
\right].
\end{equation}
Here, $\mathcal{R}$ denotes the differentiable renderer, $\phi$ the parameters of the pretrained diffusion model, and $I_t$ the noisy image at timestep $t$. The term $\hat{\epsilon}_{\phi}$ represents the predicted noise conditioned on the reference image $\tilde{I}$, while $\epsilon \sim \mathcal{N}(0, I)$ is the sampled noise and $\omega(t)$ is a timestep-dependent weighting function.

We adopt a monotonic time-annealing schedule~\cite{huang2024dreamtime}, in which the maximum noise timestep $t$ is gradually decreased during optimization. This encourages coarse, global structure formation in early iterations (large $t$), followed by refinement of fine details in later stages (small $t$).

\paragraph{Geometric Space Loss ($\mathcal{L}_{\mathbb{G}}$).}To prevent high-frequency artifacts and ensure physical realizability, we impose a smoothness regularization based on the third derivative of the curve trajectory $\mathcal{L}_{\mathbb{G}} = \mathcal{L}_{\text{jerk}}\: (\text{Eq. (2)})\:$
Minimizing the jerk energy encourages the formation of fair, aesthetically pleasing curves with minimal rate of change in curvature, mimicking the mechanical properties of a physical wire.

\begin{figure}[]
  \centering
  \includegraphics[width=\linewidth]{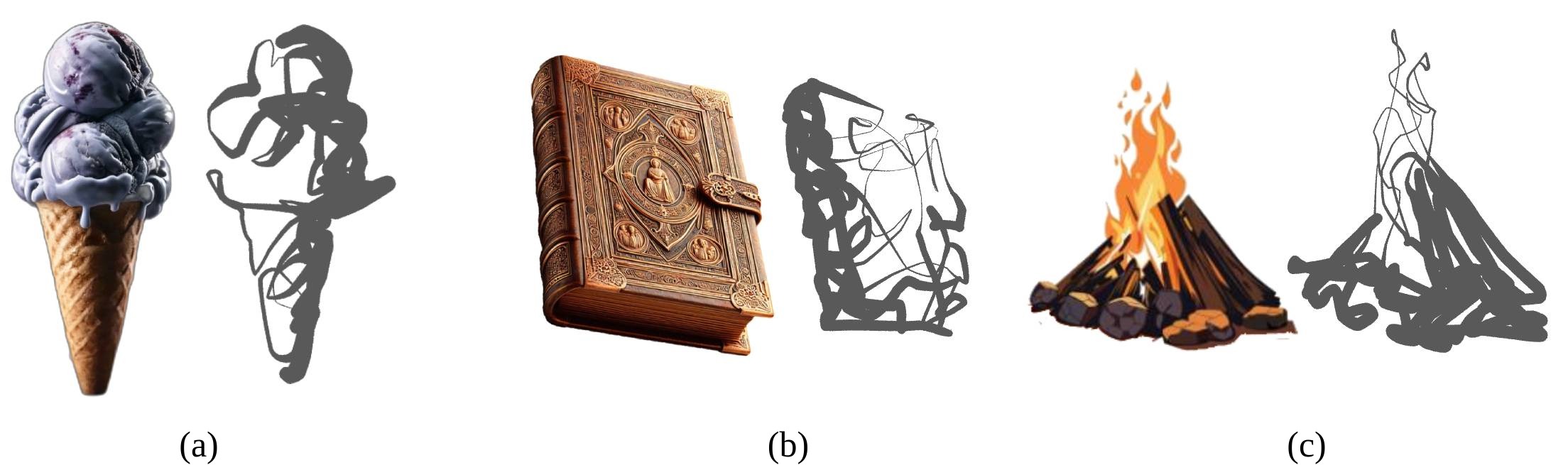}
  \caption{The 4D Wire as a Semantic Bridge. Stroke width $w(t)$ encodes more than thickness; it translates 3D attributes into 1D visual cues. (a) Lighting: Thicker strokes are utilized as a proxy for darker shading. (b) Semantic Saliency: Critical structural elements, such as a book’s spine, are prioritized via increased stroke weight. (c) Material Density: Width distinguishes between different physical states, contrasting the buoyant, amorphous nature of fire against the solid, heavy mass of the fuel logs.}
  \label{fig:encode_meaning}
\end{figure}

%%<<<<<<<<<<<
Our method achieves improved abstraction capability, as illustrated in Fig.~\ref{fig:comparison_diff3ds}. A key factor driving this performance is the introduction of a learnable width parameter $w(t)$, which allows the model to modulate the visual weight of the curve according to local semantics. For example, a single 4D Wire with varying thickness can represent structures such as multiple tree leaves within a single stroke, whereas standard constant-width Bézier primitives require dense clusters of overlapping curves to approximate similar volumetric effects. Such curve bundling introduces redundant geometry and visual clutter, increasing topological complexity without improving semantic clarity. Beyond structural abstraction, our model leverages thickness as a semantic channel, encoding attributes such as shading, hierarchical importance, and physical properties (e.g., mass or rigidity), as shown in Fig.~\ref{fig:encode_meaning}. This capability enables the model to introduce meaningful contrast along a single curve, which is difficult to achieve with uniform-width representations. Moreover, the inherent continuity of the 4D Wire encourages the optimizer to discover solutions that are both visually and structurally coherent.

\begin{figure}[]
  \centering
  \includegraphics[width=0.8\linewidth]{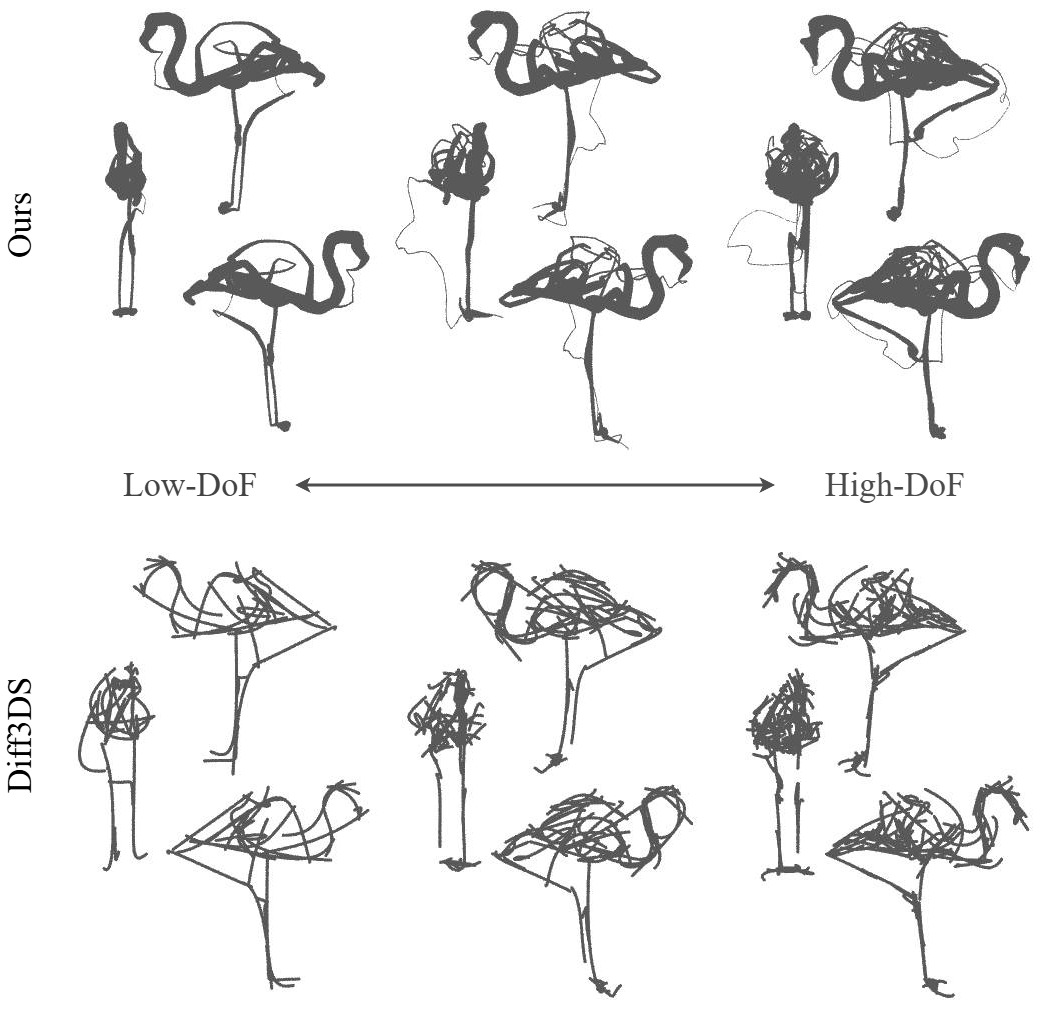}
  \caption{Effect of various degrees of freedom. from left to right with increasing keypoints.}
  \label{fig:dof}
\end{figure}

We further evaluate abstraction performance under varying degrees of freedom (DoF), controlled by the number of control points. As shown in Fig.~\ref{fig:dof}, Diff3DS~\cite{diff3ds} struggles to capture the high-curvature neck of the flamingo in low-DoF settings, reflecting the limited expressiveness of independent Bézier segments. Increasing the DoF allows Diff3DS to recover this structure, but often leads to dense clusters of overlapping strokes, resulting in visual clutter and redundant geometry. In contrast, our continuous 4D Wire scales gracefully across different DoF budgets, preserving key geometric features while maintaining a smooth and coherent topology.

\subsubsection{Quantitative Evaluation.} 
Following Diff3DS~\cite{diff3ds}, we measure semantic fidelity between the reference view and rendered novel views using CLIP~\cite{clip}, and assess reference-view structural similarity using LPIPS~\cite{lpips}. As shown in Table~\ref{tab:stroke}, our method achieves CLIP and LPIPS scores comparable to Diff3DS, despite our underlying structure being strictly connected and free of geometric clutter. This parity is expected, as both CLIP and LPIPS operate entirely in image space and are fundamentally topology-agnostic. Consequently, Diff3DS can achieve high scores by acting as an unconstrained sparse paintbrush, which often results in floating, disconnected segments that are physically impossible to fabricate. In contrast, our framework achieves the same high level of semantic and structural fidelity while strictly adhering to the rigorous topological constraint of a single, continuous wire. This proves that a single continuous curve is highly expressive, delivering superior structural integrity without sacrificing visual quality.
\begin{table}[h]
    \centering
    \caption{\textbf{Quantitative Comparison on Diff3DS.} We compare our Continuous B-Spline method against the discrete baseline (Diff3DS). Our method achieves comparable semantic accuracy (CLIP) while offering superior structural properties (Components) and aesthetic smoothness (Jerk).}
    \label{tab:stroke}
    \resizebox{\linewidth}{!}{%
    \begin{tabular}{lcccc}
        \toprule
        \textbf{Method} & \textbf{Novel-view CLIP Score} $\uparrow$ & \textbf{Reference View LPIPS} $\downarrow$\\
        \midrule
        Diff3DS \cite{diff3ds} & 0.5422 & \textbf{0.32}\\
        \textbf{Ours} & \textbf{0.5471} & 0.33\\
        \bottomrule
    \end{tabular}%
    }
\end{table}

\subsubsection{Ablation Study.}
\begin{figure}[]
  \centering
  \includegraphics[width=\linewidth]{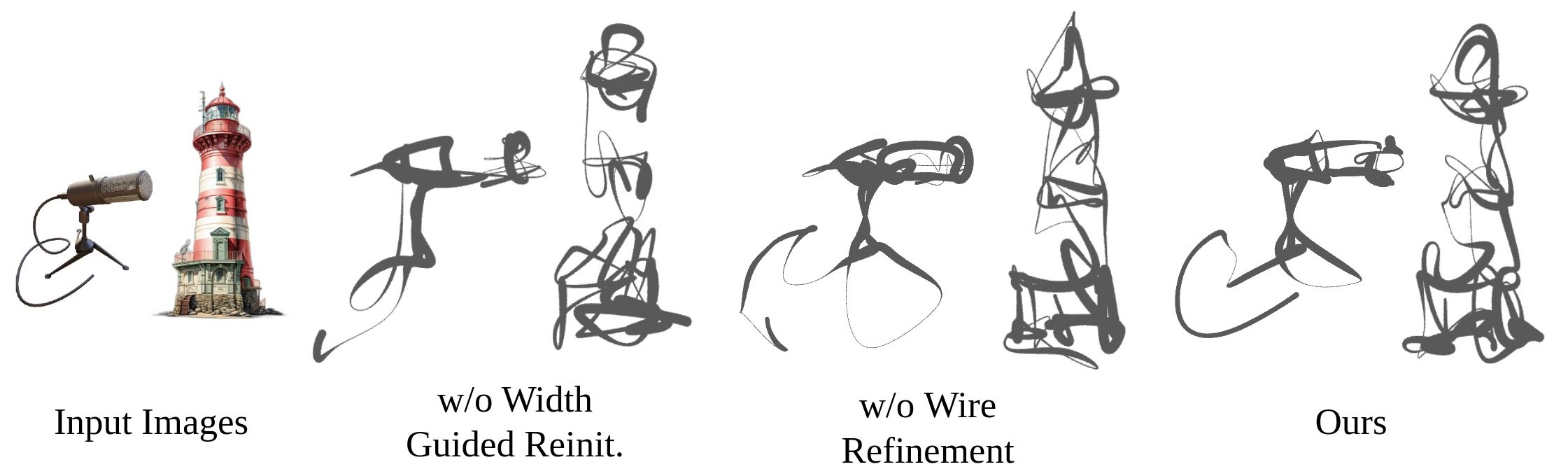}
  \caption{Ablation of key components. Removing either the width-guided reinitialization or the wire refinement module results in underfitting or reduced geometric detail, highlighting the importance of the full formulation.}
  \label{fig:stroke_ablation}
\end{figure}

We perform an ablation study to evaluate the individual contributions of \emph{width-guided reinitialization} and \emph{wire refinement} on the image-to-3D stroke task, as visualized in Fig.~\ref{fig:stroke_ablation}. Without \emph{width-guided reinitialization}, the optimization frequently converges to suboptimal local minima, yielding a curve that fails to capture the global topological structure of the target. Conversely, omitting \emph{wire refinement} limits the model's geometric resolution, degrading its capacity to represent high-frequency details, such as the wire of a microphone or the lantern (top) of a lighthouse. Together, these modules ensure both global alignment and local precision.

\subsubsection{Bad Cases.} 
\begin{figure}[]
  \centering
  \includegraphics[width=\linewidth]{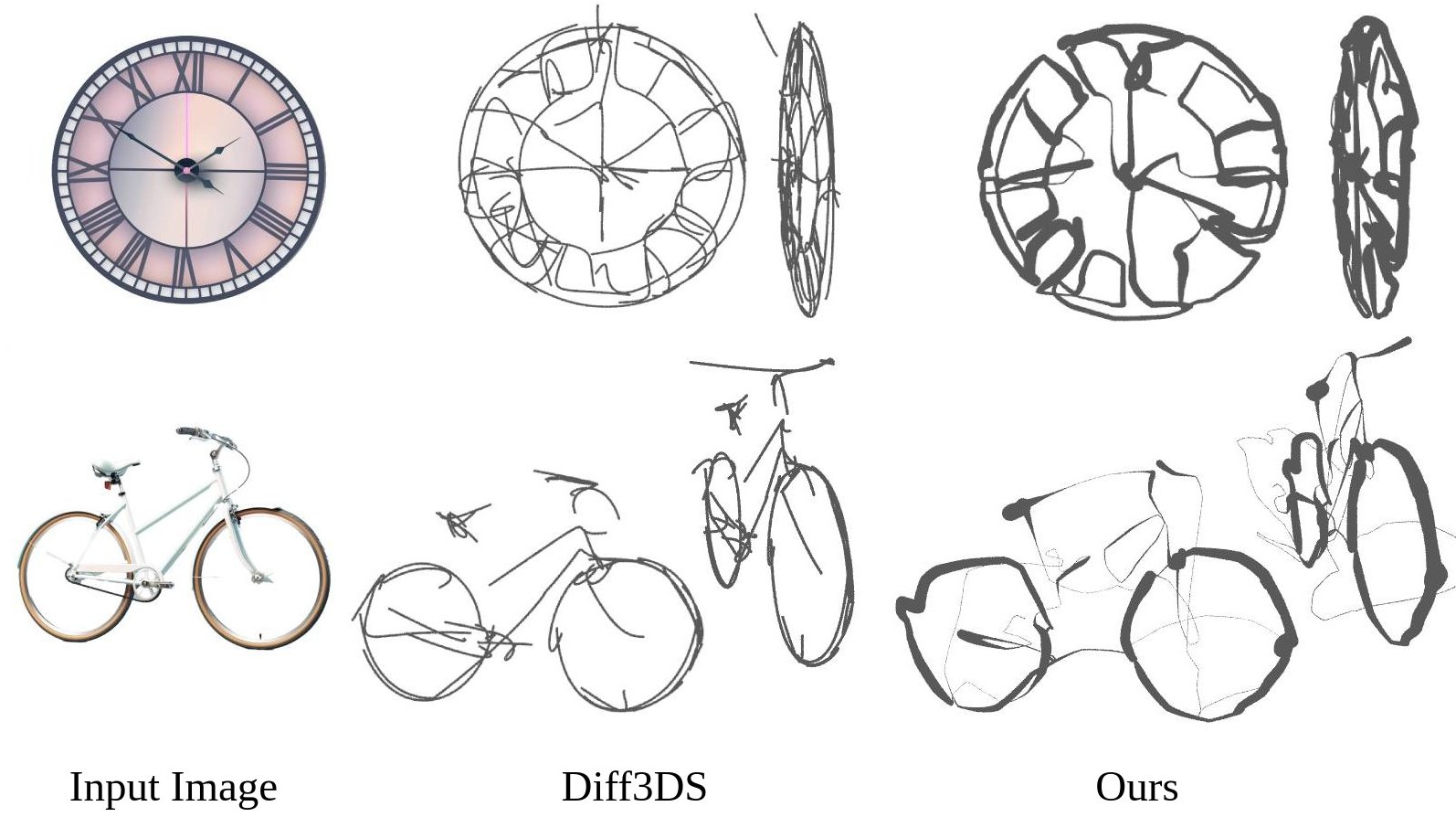}
  \caption{Our 4D wire is less effective than fixed-thickness models at capturing objects with perfectly constant thickness. The learnable width parameter introduces unnecessary degrees of freedom in circular structures like clock faces or wheels, occasionally resulting in non-uniform thickness artifacts.}
  \label{fig:basd_case}
\end{figure}

While the variable-width 4D Wire offers significantly enhanced expressivity, this flexibility introduces a minor trade-off in geometric stability. When reconstructing highly regular primitives such as the perfect circles of a clock face, the learnable width parameter $w(t)$ can struggle to maintain absolute constancy. In these specific cases, models with a hard-coded fixed thickness constraint inherently achieve uniform radial consistency, as illustrated in Fig. \ref{fig:basd_case}. This behavior is a natural consequence of the increased degrees of freedom in our representation; the optimizer may introduce subtle, non-physical width fluctuations to capture localized photometric noise or sampling artifacts. Fortunately, this is easily mitigated by restricting the $w(t)$'s variation range to a narrower interval.

\begin{figure*}[t]
  \centering
  \includegraphics[width=0.85\linewidth]{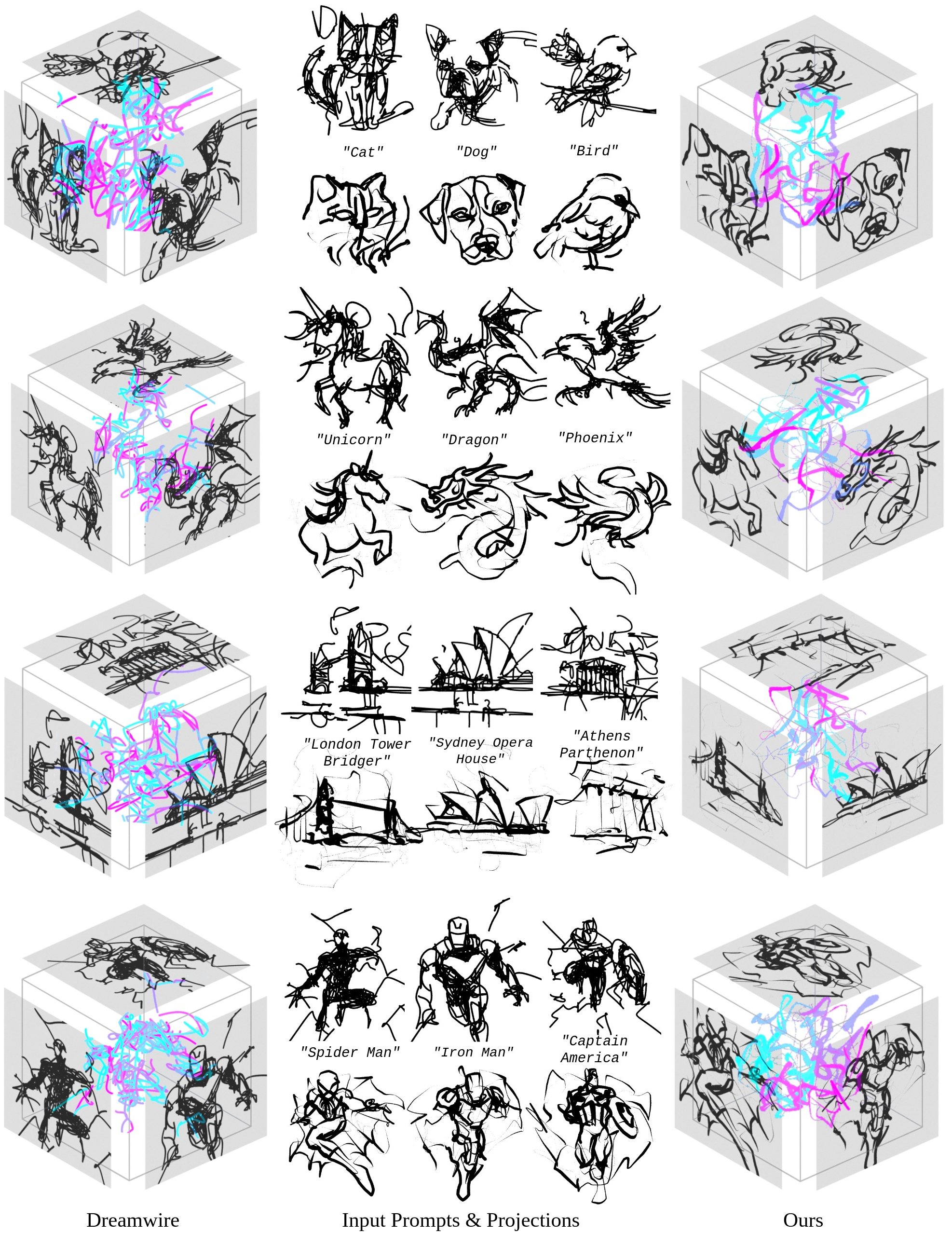}
  \caption{Given multi-view prompts (middle), we compare Dreamwire's discrete curves (left, random colors for each Bézier segment) against our 4D Wire (right, color-coded by curve parameter $t$).}
  \label{fig:comparison_dreamwire}
\end{figure*}

\subsection{Multiview Wire Art (Anamorphic Sculpting)}

%%%%%<<<<<<<<<<<<<<<<
Wire art represents 3D forms using continuous strands of material, emphasizing simplicity, continuity, and structural clarity. A defining characteristic of this medium is its ability to convey complex shapes through minimal, uninterrupted curves, producing clean and visually coherent abstractions. We investigate the application of our 4D Wire representation to Multi-View Wire Art (MVWA) generation~\cite{mvwa}, a highly constrained optimization problem in which a single structure must simultaneously depict distinct semantic content across three canonical views (front, side, and top). This setting serves as a rigorous stress test for both the expressiveness and stability of our representation. Following the framework of DreamWire~\cite{dreamwire}, we replace the baseline’s discrete Bézier chain (each composed of five segments) with a 4D Wire representation and optimize it under guidance from a pretrained 2D diffusion model.

\paragraph{Optimization.}
Since text-to-3D generation lacks explicit geometric structure, we adopt a general geometric initialization strategy to distribute points evenly inside a sphere and solve a Traveling Salesperson Problem (TSP) to obtain a single continuous path.

For image space loss $\mathcal{L}_{\mathbb{I}}$, we employ Score Distillation Sampling (SDS) with a pretrained Stable Diffusion model (SD 1.5~\cite{sd}) to supervise the optimization using three canonical views $\{(R_v, T_v)\}_{v=1}^{3}$. The resulting gradient is defined as the sum of per-view SDS gradients:
\begin{equation}
\nabla_{\tilde{\Theta}}\mathcal{L}_{\mathbb{I}} =
\sum_{v=1}^{3}
\nabla_{\tilde{\Theta}}\mathcal{L}_{\mathbb{SDS}}(\phi, I^{(v)}),
\end{equation}
where $I^{(v)} = \mathcal{R}(\tilde{\Theta}; R_v, T_v)$.

We further regularize the curve using a smoothness prior based on jerk minimization, $\mathcal{L}_{\mathbb{G}} = \mathcal{L}_{\text{jerk}}$, to encourage continuous curvature. We adopt the time-annealing schedule of~\cite{huang2024dreamtime} and set the guidance scale to 100, following DreamWire~\cite{dreamwire}. To improve optimization stability and avoid poor local minima, we incorporate \emph{width-guided reinitialization} and \emph{wire refinement} during training.

\paragraph{Results and Discussion.}
We compare our method with the baseline in Fig.~\ref{fig:comparison_dreamwire}. Despite imposing a seemingly restrictive topological constraint—a single continuous path—we observe significant improvements in both geometric coherence and semantic interpretability (as reflected in higher CLIP scores). In contrast to the ``bag of curves'' paradigm, which often relies on stochastic accumulation to satisfy multi-view constraints, our continuity constraint acts as an implicit regularizer, encouraging the discovery of globally consistent structures rather than local, degenerate solutions.

\paragraph{Stochastic Accumulation vs. Structural Routing.}
A fundamental distinction between our approach and prior work such as DreamWire lies in how the model reasons about the curve. As shown in Fig.~\ref{fig:comparison_dreamwire}, DreamWire constructs images through the accumulation of many independent strokes. In this regime, individual primitives act as local texture elements, contributing limited semantic meaning in isolation. The model can satisfy visual objectives by increasing stroke density in specific regions, effectively bypassing global geometric reasoning. In contrast, our continuous representation transforms the problem into a structural routing task. Rather than introducing independent primitives, the optimizer must determine how a single curve traverses the entire shape. As a result, the method shifts from local appearance matching toward true geometric abstraction. This behavior is evident in our results, where the learned wire aligns with meaningful contour structures and acts as a critica; geometric element across all three views.

\begin{figure}[t]
  \centering
  \includegraphics[width=\linewidth]{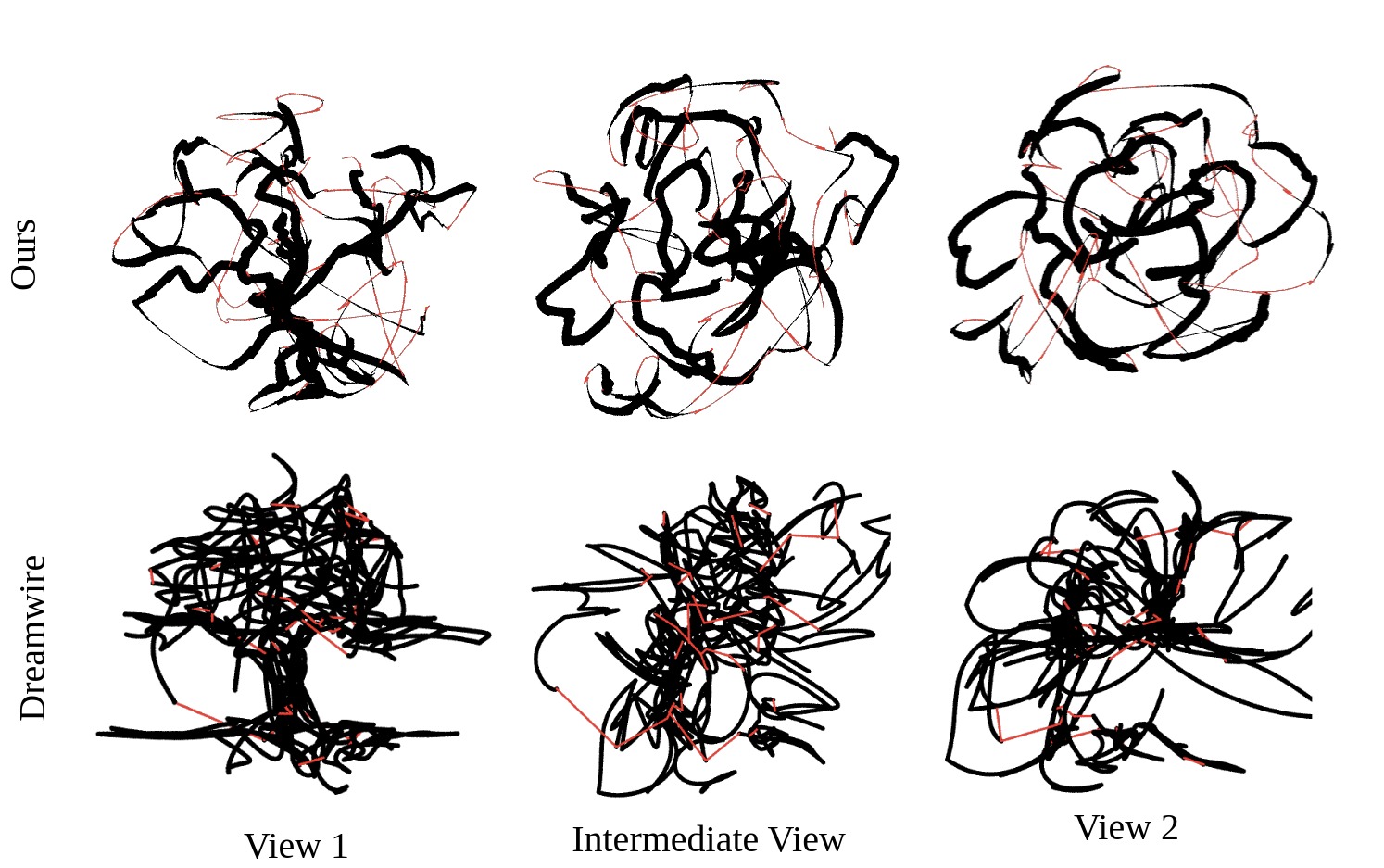}
  \caption{Connection visualization.}
  \label{fig:connection}
\end{figure}

\paragraph{Connectivity and Gestalt Coherence.}
Finally, we consider the relationship between physical and perceptual connectivity. DreamWire optimizes for proximity between strokes but does not guarantee mathematical continuity, often requiring post-processing (e.g., Minimum Spanning Tree construction) to form a connected structure (Fig.~\ref{fig:connection}). In contrast, our 4D Wire is inherently continuous. Although the learnable width $w(t)$ allows portions of the curve to become visually thin or nearly invisible, the underlying trajectory remains continuous. This aligns with the Gestalt law of continuity~\cite{palmer1999vision}, where elements arranged along smooth, continuous paths are perceptually grouped as a single structure. This perceptual coherence contributes to the improved visual quality and consistency of our results.

%>>>>>>>>>>>>>>>>>>>

\begin{figure}[t]
  \centering
  \includegraphics[width=\linewidth]{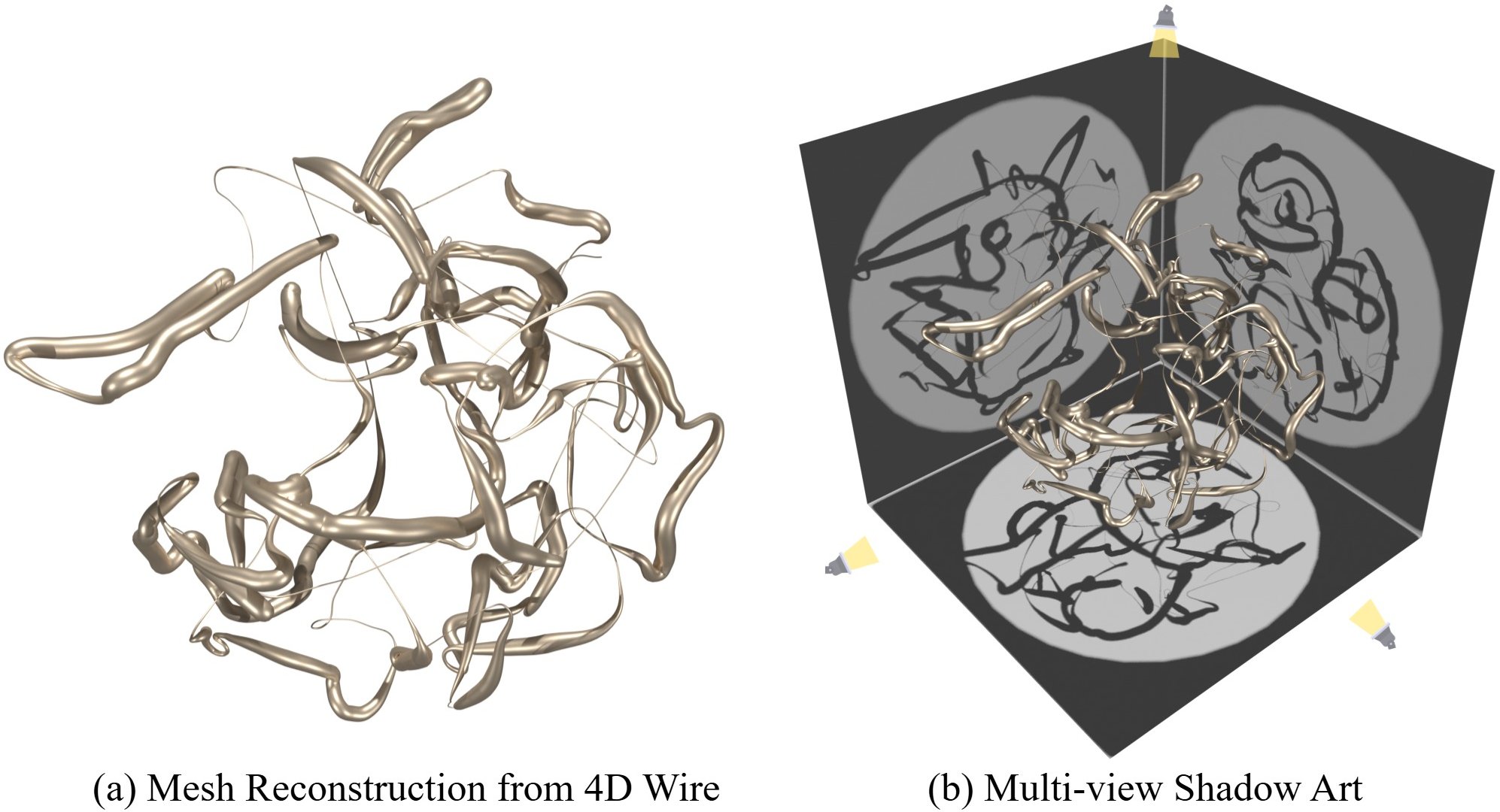}
  
  \caption{(a) Our 4D Wire is seamlessly translated into a mesh-based format for advanced rendering. (b) A Shadow Art application, utilizing light-path tracing to project light onto the sculpture and cast view-dependent shadows.}
  \label{fig:shadow_art}
\end{figure}

Furthermore, our 4D B-spline representation is inherently compatible with industry-standard mesh pipelines, enabling seamless integration into downstream workflows such as manual sculpting, high-fidelity texturing, and production-grade rendering. We demonstrate this interoperability through a \emph{shadow art} application (Fig.~\ref{fig:shadow_art}). Specifically, we convert the optimized 4D Wire into a manifold mesh and perform physically-based light transport simulation, allowing the multi-view wire structure to be realized as a shadow art installation.

\subsubsection{Quantitative Evaluation}

To evaluate our 4D wire framework against the baseline method, DreamWire \cite{dreamwire}, we employ a suite of metrics assessing both semantic fidelity and structural characteristic on 32 cases with categories extracted from the QuickDraw dataset \cite{quickdraw}:

\begin{itemize}
\item \textbf{Semantic Consistency (CLIP):} Similar to DreamWire, we compute the text-image consistency of the rendered views with the input text prompt averaged across three views with CLIP similarity \cite{clip}.
\item \textbf{Structural Integrity (\# Components):} This metric counts the number of disconnected components in the generated mesh. Ideally, $N_{comp}=1$. A lower count indicates superior physical connectivity.
\item \textbf{Total Length:} We measure the total length of the curve used. This serves as a proxy for abstraction simplicity.
\item \textbf{Connectivity Cost:} To quantify the magnitude of fragmentation in the baseline (DreamWire), we calculate the Minimum Spanning Tree (MST) weight required to bridge all discrete components.
\end{itemize}

\begin{table}[h]
    \centering
    \caption{\textbf{Quantitative Comparison on Multiview Wire Art.}}
    \label{tab:comparison}
    \resizebox{\linewidth}{!}{%
    \begin{tabular}{lcccc}
        \toprule
        \textbf{Method} & \textbf{CLIP Score} $\uparrow$ & \textbf{\# Components} $\downarrow$ & \textbf{Length} $\downarrow$ & \textbf{Connectivity Cost} $\downarrow$ \\
        \midrule
        DreamWire \cite{dreamwire} & 0.307 & 38& 116.9 & 7.5 \\
        \textbf{Ours} & \textbf{0.323} & \textbf{19} & \textbf{40.18} & \textbf{3.1} \\
        \bottomrule
    \end{tabular}%
    }
\end{table}

As detailed in Table~\ref{tab:comparison}, our method consistently achieves higher CLIP scores, indicating closer alignment between the rendering and the input text prompts compared to the baseline (DreamWire). Notably, this improved semantic fidelity is obtained with a significantly shorter total curve length. Together, these results suggest that our method achieves a higher semantic density, conveying more meaningful visual information per unit of geometry.

In terms of structural properties, DreamWire produces fundamentally disconnected representations, resulting in a large number of connected components. Although our representation is intrinsically topologically continuous, we adopt a conservative evaluation protocol for fair comparison: segments with near-zero width (introduced for occlusion handling) are treated as structural breaks. Even under this strict criterion, the baseline exhibits a higher number of components and connectivity cost, indicating a greater degree of fragmentation.

\subsubsection{Ablation Study}
\begin{figure}[t]
  \centering
  \includegraphics[width=\linewidth]{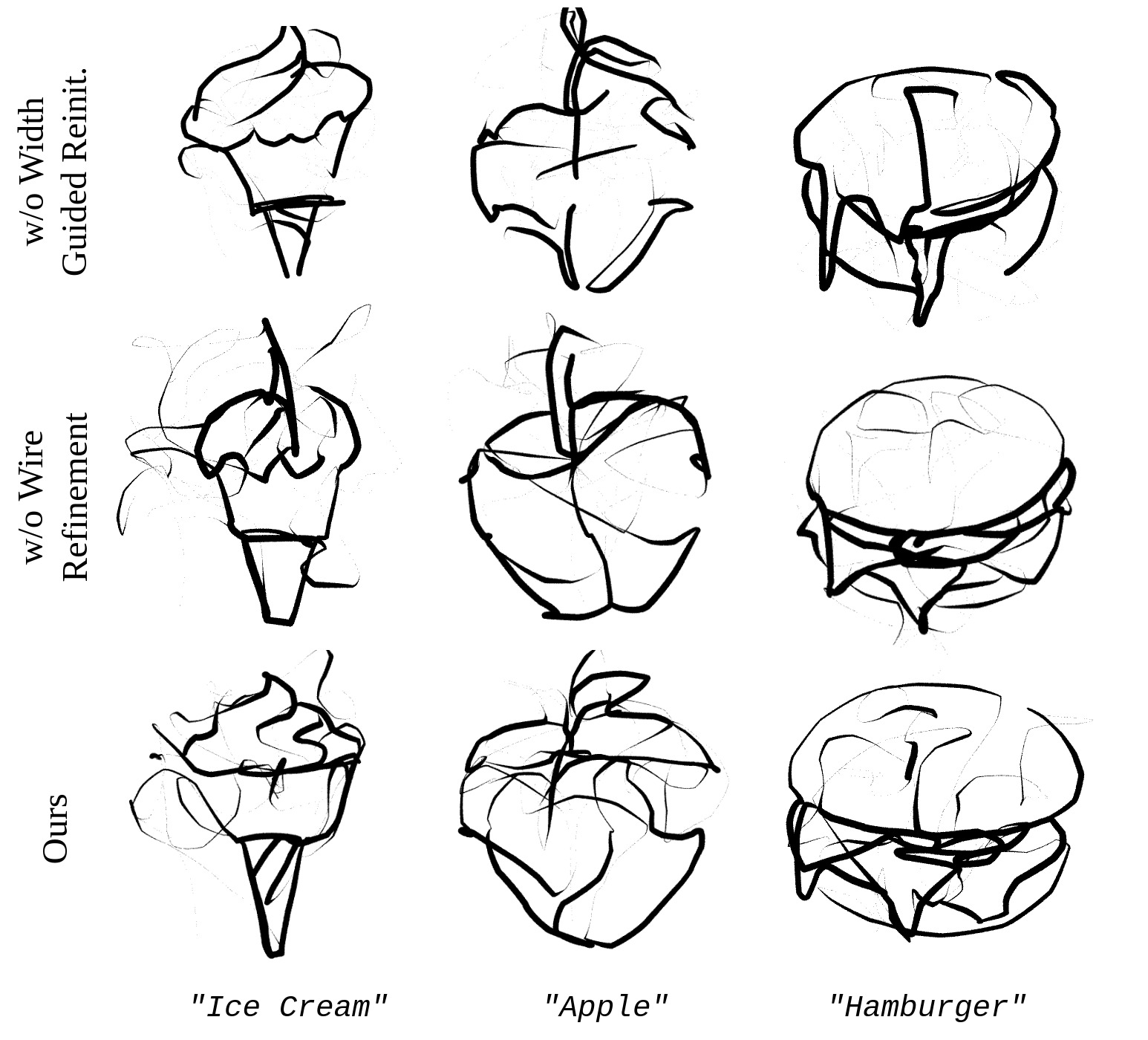}
  \caption{Ablation of key components. Removing either the width-guided reinitialization or the wire refinement module results in underfitting or reduced geometric detail, highlighting the importance of the full formulation.}
  \label{fig:mvwa_ablation}
\end{figure}

We conduct an ablation study to evaluate the individual contributions of \emph{width-guided reinitialization} and \emph{wire refinement} (Fig.~\ref{fig:mvwa_ablation}). Our results indicate that \emph{width-guided reinitialization} is critical for regularizing the optimization landscape, preventing the model from collapsing into suboptimal local minima. Conversely, omitting \emph{wire refinement} restricts the available geometric degrees of freedom, hindering the reconstruction of high-frequency details such as the layered textures of the ice cream cone and hamburger.

\subsubsection{Bad Case Analysis}
\begin{figure}[t]
  \centering
  \includegraphics[width=\linewidth]{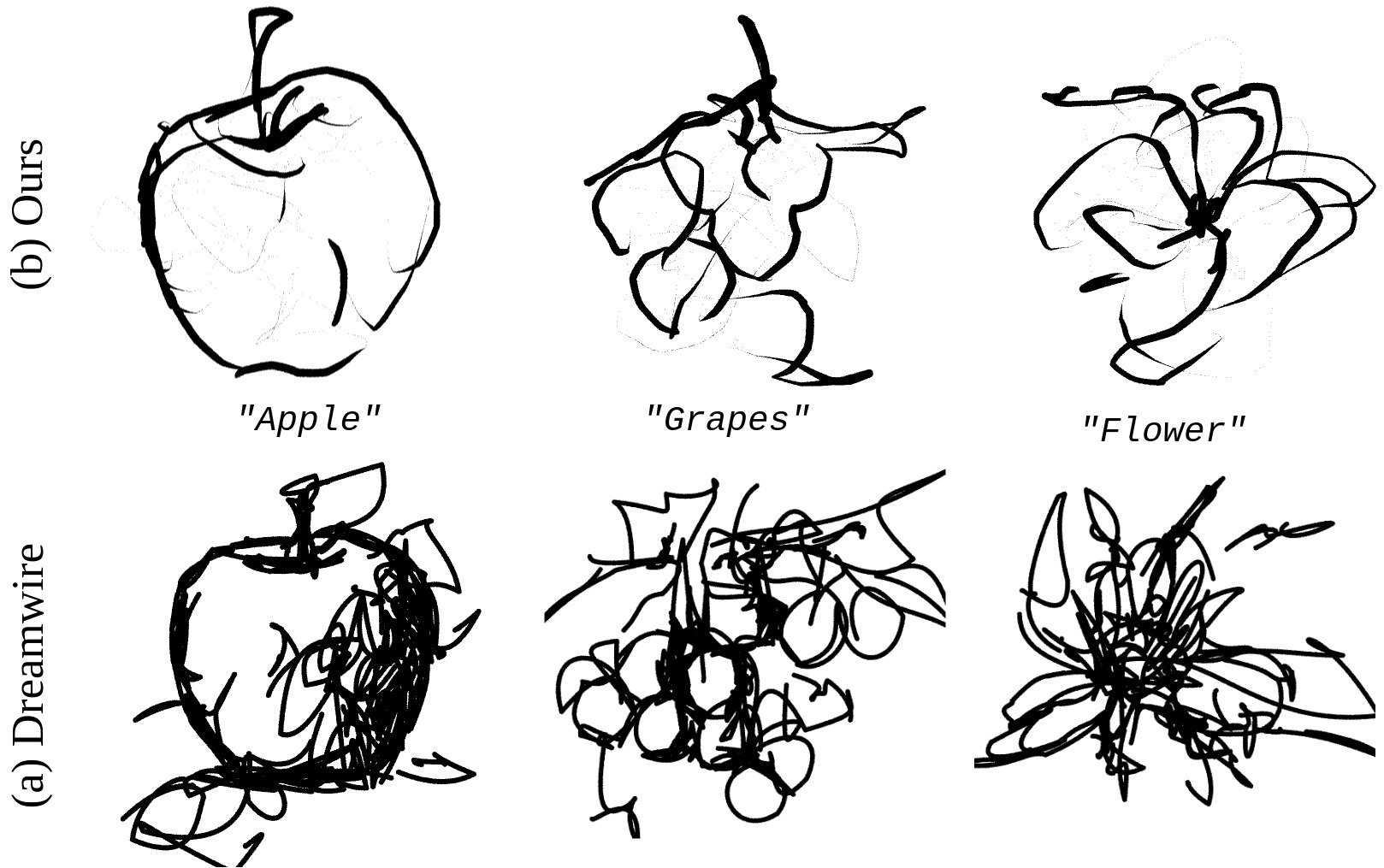}
  \caption{Failure case in multi-view wire art. Inconsistent prompts across views lead to conflicting geometric constraints, resulting in degraded result in some view.}
  \label{fig:mvwa_bad}
\end{figure}

Similar to DreamWire, our method is affected by spatial inconsistencies across multi-view prompts. For example, one view may correspond to a relatively simple structure (e.g., an apple), while another requires a more complex arrangement (e.g., grapes), or different views may impose conflicting aspect ratios (e.g., tall versus wide shapes). Such inconsistencies can lead to configurations that are difficult—or impossible—to satisfy simultaneously with a single continuous structure. In contrast to DreamWire, our 4D Wire representation exhibits an inherent simplicity bias. When faced with conflicting constraints, the optimization tends to favor simpler geometric configurations, often sacrificing fine details required by more complex views. As shown in Fig.~\ref{fig:mvwa_bad}, our method prioritizes the simpler apple view while reducing the structural complexity of the grape view. This behavior arises from the variable-width formulation, which allows the curve thickness to shrink toward zero in regions of conflict, effectively pruning incompatible geometry. In contrast, DreamWire operates on a fixed set of curve primitives that cannot be removed during optimization, and thus tends to preserve complexity in more demanding views at the expense of simpler ones, often resulting in redundant geometry (e.g., excessive strokes in the apple view).

\subsection{Differentiable Stylized Surface-Filling}
\begin{figure}[h]
  \centering
  \includegraphics[width=\linewidth]{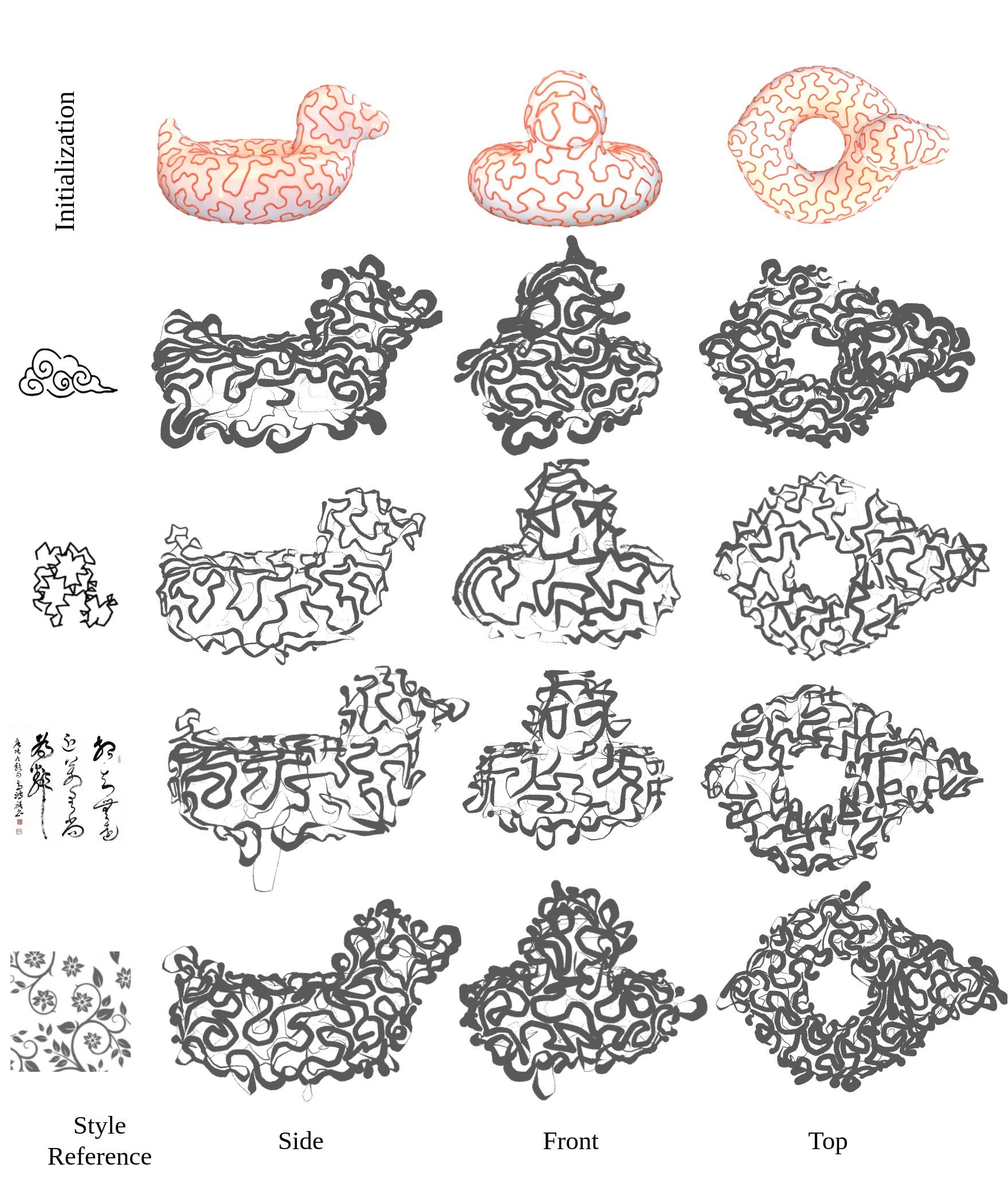}
  \caption{Stylized Surface Filling. Top: Input mesh and initialization \cite{sfc2024}. Left: The target style reference.}
  \label{fig:semantic_sfc}
\end{figure}

Space-filling curves are a fundamental concept in computational geometry, serving as a bridge between 1D sequences and higher-dimensional domains \cite{Hilbert1891, Peano1890}. While originally theoretical, they have found critical practical applications in 3D printing infill patterns, CNC toolpath planning, and procedural art \cite{maze}. However, existing approaches for 3D surface filling remain predominantly functional—optimizing strictly for geometric coverage or fabrication constraints—rather than expressive. We observe a rich body of work in 2D aesthetic filling, where curves are optimized to match silhouettes or pack object within an area \cite{2dbspline, image_collage}. In this section, we propose a framework for stylized surface filling, building on upon the idea in 2D. We transition from purely geometric coverage to semantically-driven aesthetic filling, enabling continuous curves that not only cover a 3D shape but also embody semantic concepts like "calligraphy or "flower" (Fig. \ref{fig:semantic_sfc}) through differentiable optimization.

\paragraph{Initialization.}
To establish a topologically valid initialization, we first generate a continuous surface-filling curve using the method of~\cite{sfc2024} on the input mesh. The key parameter controlling this process is the fill density, defined by the inter-curve spacing. This parameter introduces a trade-off: higher density improves geometric fidelity, while lower density encourages abstraction and enhances stroke clarity. In practice, we select a moderate density (spacing $= 0.06$ for objects normalized to a unit sphere) that balances coverage and visual simplicity, preserving the overall silhouette while avoiding excessive clutter. The resulting discrete trajectory is then fitted with our continuous 4D B-spline representation, initialized with uniform thickness.

\paragraph{Optimization.}
Our optimization objective balances two competing goals: uniform surface coverage and semantic stylization. To achieve aesthetic coverage on a curved manifold, we propose a Multi-view MMSE (Multi-scale Mean Squared Error \cite{2dbspline, image_collage}) loss. Our loss aggregates MMSE against the object silhouette ($Mask_v$) across multiple randomized viewpoints $v \in \mathcal{V}$, ensuring the curve is aesthetically pleasing from all angles. 

To inject semantic style, we adopt the patch-based Directional CLIP loss \cite{Kwon_2022_CVPR} via CLIP-AG \cite{clipag}. Rather than minimizing absolute distance—which often causes adversarial noise in vector optimization—this method aligns the semantic trajectory of local rendered patches with a target stylistic direction $\Delta S$ in the CLIP latent space. The global direction $\Delta S$ is pre-computed from either paired text prompts (e.g., "a continuous 3D wire" $\to$ "a flower") or reference images. During optimization, we compute the normalized latent direction $\Delta I_p$ between a given rendered patch and its unstyled initialization. The stylistic alignment maximizes their cosine similarity:
\begin{equation}
\mathcal{L}_{\text{CLIP}} = 1 - \frac{1}{|\mathcal{P}|} \sum_{p \in \mathcal{P}} \cos(\Delta I_p, \Delta S),
\end{equation}
where $\mathcal{P}$ represents the set of locally sampled image patches. 

\begin{figure}[h]
  \centering
  \includegraphics[width=\linewidth]{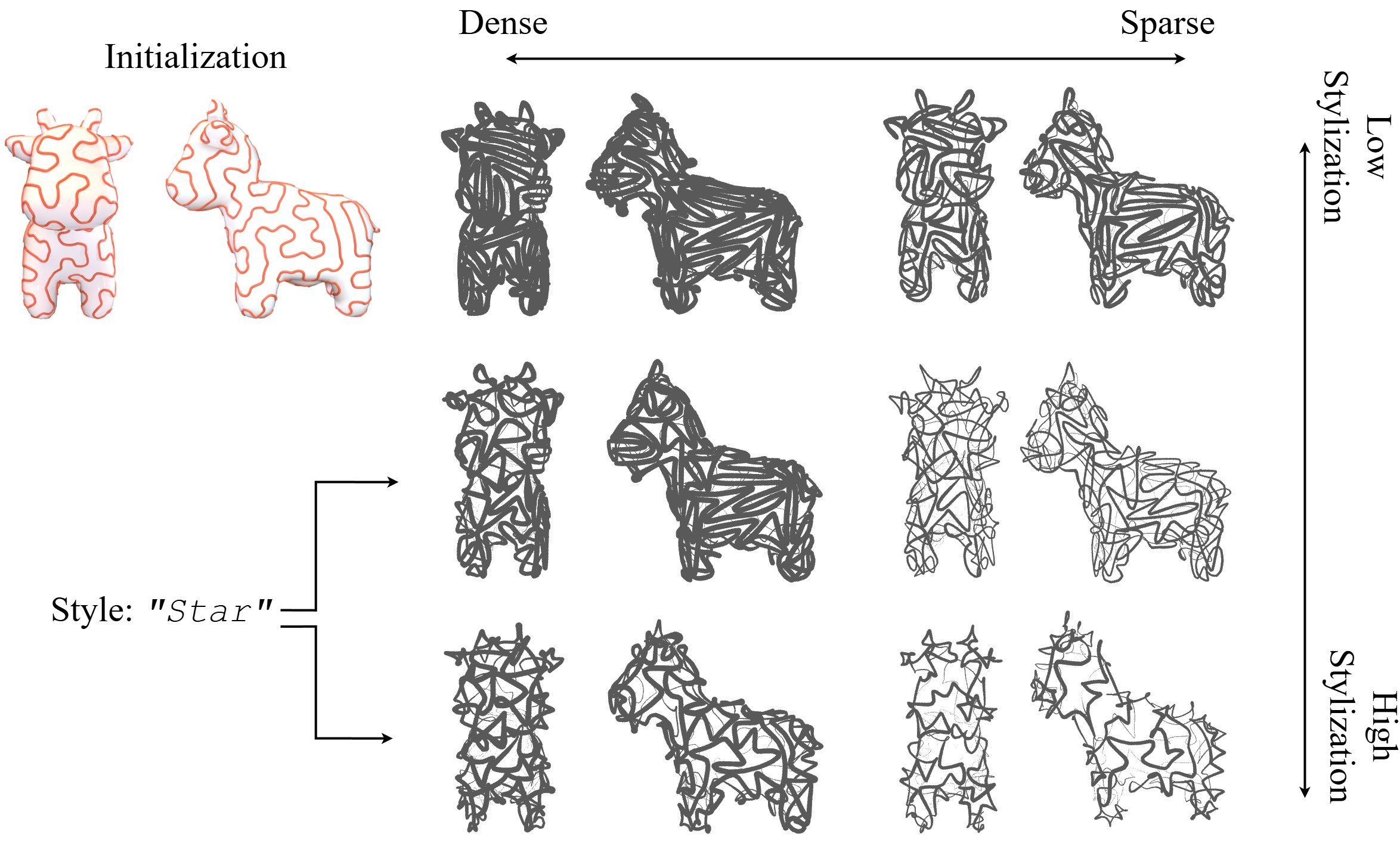}
  \caption{Configurable Surface Filling. The input 3D mesh is initialized using the geodesic curve generation of \cite{sfc2024} (shown in orange). Our framework offers granular control over fill density (ranging from sparse to dense) and the magnitude of stylization, enabling results that balance abstract geometric coverage with semantic expression.}
  \label{fig:sfc}
\end{figure}

The total image-space loss is defined as:
\begin{equation}
\mathcal{L}_{\mathbb{I}} = \sum_{v \in \mathcal{V}} \Big( \lambda_{\text{mmse}} \mathcal{L}_{\text{MMSE}}(I_v, \alpha \cdot Mask_v) + \lambda_{\text{clip}} \mathcal{L}_{\text{CLIP}} \Big)
\end{equation}
where $\alpha \in (0, 1]$ is a configurable opacity constant applied to the target silhouette mask. By modulating $\alpha$, we directly control the spatial density of the optimized curve: a lower $\alpha$ reduces the target pixel intensity, yielding a lighter wireframe, while a higher $\alpha$ forces the curve to bunch densely to match a solid silhouette, as demonstrated along the density axis in Fig. \ref{fig:sfc}. 

Furthermore, the hyperparameters $\lambda_{\text{mmse}}$ and $\lambda_{\text{clip}}$ govern the inherent trade-off between this uniform spatial distribution and semantic deformation (e.g., locally bending or bunching the curve to form specific visual features), which is illustrated along the stylization axis in Fig. \ref{fig:sfc}. Finally, to prevent structural degradation during this process, we maintain a geometric regularization term, $\mathcal{L}_{\mathbb{G}} = \mathcal{L}_{\text{jerk}}$, ensuring the curve retains smooth physical properties throughout the optimization.

\paragraph{View-Dependent Visibility} 
A primary challenge in optimizing surface-filling curves is \emph{wireframe clutter} caused by back-facing geometry. When occluded segments are rendered, the 2D diffusion prior observes a visually tangled projection in which both front- and back-facing structures are simultaneously visible, whereas the optimization should focus only on the visible surface. A straightforward solution is to mask occluded regions during rendering. However, standard \emph{hard} depth clipping—where hidden segments are geometrically truncated or discarded—introduces abrupt visual transitions at occlusion boundaries and complicates implementation due to dynamically changing topology. To address this issue, we introduce a differentiable soft visibility mechanism based on backside attenuation. During rendering, the projected depth of the curve, $z_{\text{curve}}$, is continuously compared against the mesh depth buffer, $z_{\text{mesh}}$. Instead of performing discrete occlusion culling, we compute a continuous visibility score $V$ and use it to smoothly attenuate the local stroke width $w(t)$ toward zero:

\begin{equation}
\begin{aligned}
    &V = \sigma\big(k \cdot (z_{\text{mesh}} - z_{\text{curve}} + b)\big) \\
    &w_{\text{final}}(t) = w_{\text{base}}(t) \cdot V
\end{aligned}
\end{equation}
where $\sigma$ is the sigmoid activation function, $k$ controls the sharpness of the transition, and $b$ is a shift bias (e.g., $b=0.05$). This bias shifts the logistic function to ensure that curves resting perfectly on the mesh surface ($z_{\text{mesh}} = z_{\text{curve}}$) evaluate to near-full visibility. 

\section{Conclusion}
In this work, we introduce the 4D Wire primitive, a continuous B-spline framework that addresses the structural limitations of the "bag-of-curves" paradigm. By optimizing a single trajectory with variable width $(x, y, z, w)$, we show that enforcing topological continuity provides a strong inductive bias, transforming the generation process from local stochastic accumulation to global semantic routing. Our approach produces results that are both visually coherent and structurally connected, as demonstrated across image-to-3D stroke, multi-view wire art, and stylized surface filling tasks. Future work includes incorporating fabrication-aware constraints to further bridge the gap between computational abstraction and physical realization.

\bibliographystyle{unsrtnat}
\bibliography{references}  %%% Uncomment this line and comment out the ``thebibliography'' section below to use the external .bib file (using bibtex) .





% \end{thebibliography}

\end{document}